\documentclass[runningheads]{llncs}

\usepackage{url}
\usepackage[hidelinks]{hyperref}
\usepackage[utf8]{inputenc}
\usepackage{graphicx}
\usepackage[caption=false]{subfig}
\usepackage{amsmath}
\usepackage{amssymb}

\usepackage{booktabs}
\usepackage{tabularx}
\usepackage{todonotes}
\usepackage{microtype}

\providecommand{\citet}{}
\renewcommand{\citet}[1]{\citeauthor{#1} (\citeyear{#1})}
\def\eg{\textit{e.g.}}
\def\ie{\textit{i.e.}}

\title{Robustness of Explanation Methods for NLP Models}

\author{Shriya Atmakuri\inst{1}\thanks{The first three authors have equal contribution.}\and
Tejas Chheda\inst{1}\and
Dinesh Kandula\inst{1}\and
Nishant	Yadav\inst{1}\and
Taesung~Lee\inst{2}\and
Hessel Tuinhof\inst{2}
}
\authorrunning{S. Atmakuri et al.}
\institute{University of Massachusetts Amherst \\ Manning College of Information and Computer Sciences \and
 IBM Research}

\begin{document}

\maketitle

\begin{abstract}
Explanation methods have emerged as an important tool to highlight the features responsible for the predictions of neural networks. There is mounting evidence that many explanation methods are rather unreliable and susceptible to malicious manipulations. In this paper, we particularly aim to understand the robustness of explanation methods in the context of text modality. We provide initial insights and results towards devising a successful adversarial attack against text explanations.
To our knowledge, this is the first attempt to evaluate the adversarial robustness of an explanation method.
Our experiments show the explanation method can be
largely disturbed for up to 86\% of the tested samples
with small changes in the input sentence and its semantics.

\end{abstract}

\section{Introduction}
Large and complex neural network models have become state-of-the-art in many computer vision and natural language processing tasks. However, the complexity that results in their effectiveness also causes a lack of interpretability. This is a major disadvantage of these models and makes it difficult to deploy them in sensitive applications where `black box' solutions do not suffice. To combat this, a number of explainability methods have been developed.
As deep neural networks (DNN) are being deployed in critical fields like autonomous driving and healthcare, explanations can help satisfy regulatory requirements \cite{Goodman}, detect adversarial inputs, help practitioners debug their model, and reveal bias or other unintended effects learned by a model.

Intensive research on improving the DNN explainability has resulted in several either model-level or instance-level explanation methods. Prominent among these are gradient-based methods such as saliency mapping. As these methods are widely adopted, so too does the need to ensure that they behave in a reliable manner. Unfortunately, recent research has shed doubt on the validity and exposed vulnerability of explanation methods \cite{AdebayoSanityChecks,adebayo2021post,alvarez2018robustness,dombrowski2019explanations,ghorbani2019interpretation,IDLF,zhou2021feature}. The latter work only considers continuous inputs such as image data. Similarly, work on improving the robustness of explanation methods (\cite{dombrowski2022towards,etmann2019connection,mishra2021survey}) often focuses solely on continuous inputs.

In this work, the adversarial robustness of an explanation for discrete input data will be evaluated exemplified by an NLP model, taking a saliency mapping method as an example.
% In the future revision, we need a representative example here..
In particular, we analyze how saliency maps change
when inputs (\eg, ``a gorgeous, high-spirited musical'') are perturbed to create new inputs (\eg, ``a resplendent, high-spirited musical'') that maintain
the semantics and the model prediction (\eg, positive sentiment).
We expect robust explanations to be invariant to such perturbations. Most of the prior work on evaluating the scope and quality of explanation only considers the image domain, which is continuous, and thus cannot be directly applied to text data with its discrete input nature. To the best of our knowledge, ours is the first attempt to understand the robustness of explanation methods towards adversarial text explanations in the NLP domain. We therefore hope that our work makes a first contribution towards improving the robustness of explanations in the case of discrete inputs such as text and tabular data.

Our preliminary experiments show the saliency mapping method is vulnerable to such an attack where up to 86\% of the tested samples with small input perturbations have significant shift in the saliency map. We consider four different transformations (misspellings, word deletion, synonym substitution, and word inflection), and show the robustness of the saliency mapping for varying criteria to cover different application scenarios.
\section{Related Work}
% \subsection{Explanation Methods}

In this paper, we consider 
an explanation method that receives a model and an input, and maps them to an attribution object of the same size as the input.
Each entry in the output of the explanation method describes the relevance of the corresponding entry in the input for predicting the class.

% \textbf{Gradient} method is the simplest among the explanation methods. The gradient of the output is calculated with respect to the input and each gradient serves as the importance of the corresponding feature in input.
% \textbf{Gradient}~$\bigodot$~\textbf{Input} is an extension of the gradient method. It takes gradients of the output with respect to input and multiplies them by the input feature values. An intuition for this method can be obtained by taking a linear model. The gradients are simply the coefficients of each input, and the product of the input with a coefficient corresponds to the total contribution of the feature to the linear model's output.
For example, Integrated Gradients (IG) ~\cite{sundararajan2017axiomatic} uses the gradient of the output to compute the importance of the input.
% gradient saturation by summing over scaled versions of the input. IG for an input x is defined as \\$E_{IG}(x) = (x-\overline{x}) \times \int\limits_0^1\frac{\partial S(\overline{x}+\alpha(x-\overline{x}))}{\partial x}d\alpha$
% \\where $\overline{x}$ is a baseline input.
SmoothGrad (SG)~\cite{smoothgrad} seeks to alleviate noise diffusion for saliency maps by averaging over explanations of noisy copies of an input.
% This can be used on top of any of the attribution methods.

% \subsection{Adversarial Robustness}

Past work on understanding the adversarial robustness of explanation methods focuses on image data. \cite{AdebayoSanityChecks} proposes an actionable methodology to evaluate explanations. They rely on visual information to support their findings. They find that Guided BackProp and Guided GradCAM are invariant to higher layer parameters.
Nevertheless, the paper does not try to understand the explanation methods by perturbing the input features. They instead change the model parameters and input labels.

Authors in \cite{Unreliability} introduce an input invariance axiom and propose that the axiom needs to be satisfied by explanation methods to ensure reliable explanation of the input’s contribution to the model prediction. Although the work deals with transformations of the input, it only experiments with image data. Also the paper limits itself to only one simple transformation of the input: a constant shift of the input. The constant shift transformation cannot be directly used on text inputs and we devise techniques to attack explanation methods using various transformations of the inputs.
%In their work evaluating explanation methods, \citet{yeh2019fidelity} propose metrics of infidelity and sensitivity, which are relevant to our work.
% Taesung: I commented this out as I think we don't consider this in our methods or experiments, and we do not provide the reasoning in choosing some metrics over these.

Authors in \cite{IDLF} propose a new class of attacks that generate adversarial inputs not only misleading a target DNN but also deceiving its coupled interpreter. Work of \cite{Madry} also proposed a similar attack method that generates an adversarial example using projected gradient descent. Both methods generate adversarial examples by perturbing the input which cannot be directly translated into the text domain as the perturbed embedding may not map to any word.
Moreover, we aim to maintain the model prediction to focus on the robustness of the explanation method, not as a side-effect of deceiving the model.

% We also plan to explore how the attributions\todo{Starting here...} by explanation methods are affected when adversarial training is incorporated. Typically adversarial perturbations consist of making small modifications to many real-valued inputs. Since for text-related tasks, the input is discrete, traditional adversarial training techniques used for image domain cannot be directly translated for text data. \citealt{Adversarial} performs perturbation on continuous word embeddings instead of discrete word inputs as a means of regularizing a text classifier. Their approach achieved state of the art performance for multiple semi-supervised text classification tasks, including sentiment classification and topic classification. \citealt{TIATNM} recently proposed a simple and improved vanilla adversarial training process for NLP models which is a new and cheaper word substitution attack optimized for vanilla adversarial training. TextAttack (\citealt{TextAttack}), a Python framework, has implementations of several adverserial text generation methods and can be used for adversarial training. We are planning to look into different adversarial text generation methods and record how the explanations change after incorporating adversarial training into the model.\todo[inline]{Hessel: I propose we completely remove this part for the workshop as we haven't worked on any experiments yet. If you agree with me, please someone just delete this paragraph.}

\section{Methods}
% In this section, we formally define the components we use to evaluate the adversarial robustness of an explanation method.

% \subsection{Explanation Methods}
% \input{section3/explanation_methods}
\begin{figure*}[t!]
    \centering
    \includegraphics[width=0.95\textwidth]{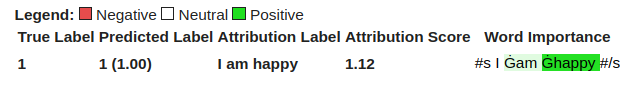}
    \caption{Input attribution generated using Captum (IG). IG assigns very high attribution score to the word {\it happy} for positive sentiment.}
    \label{fig:captum example1}
\end{figure*}
\begin{figure*}[t!]
    \centering
    \includegraphics[width=0.95\textwidth]{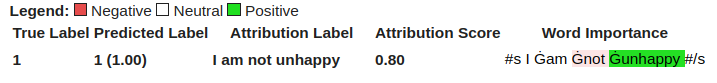}
    \caption{The word {\it not} is semantically important but is given very little weight by IG.}
    \label{fig:captum example2}
\end{figure*}

\subsection{Overview} %\todo{This is repeated in experiments? Should we keep all of this? I think we only included because proposal format required it.}

To evaluate the adversarial robustness of an NLP model, we generate adversarial examples using a saliency-based explanation method, an access to the model needed by the explanation method (often white-box for gradient-based approaches),
and transformations.
We consider a text classifier such as sentiment classification and apply perturbations to change the input sentence
such as synonym substitution, word deletion, and misspelling.
Such transformations should result in perturbed inputs that are semantically equivalent to the original input and indistinguishable to a human observer.
The transformations should also not alter the output of the model itself to reduce confounding factors. 
We then measure the change in the saliency mapping, which provides the attribution score, which show a positive or negative contribution to the predicted class.
For example, in Figure~\ref{fig:captum example1}, an explanation method assigns very high attribution score to the word {\it happy}, whereas in Figure~\ref{fig:captum example2}, when the sentence is rephrased, the word {\it not} is given negative attribution while it is very important to model's prediction of positive sentiment.

Formally, for an input $X$ consisting of a sequence of words $x_1, x_2,\ldots,x_n$ and a model $M$ that produces a prediction $P$, the explanation $E$ is defined as the vector of attributions $e_1, e_2,...,e_n$ produced by applying an explanation method $S$ to $X$ and $M$ where $e_i$ indicates the importance of the word to $M$ in producing $P$. A positive value of $e_i$ indicates that $x_i$ supported the prediction while a negative value indicates that $x_i$ contradicted the prediction.

We then apply one or more transformations $t_1,\ldots, t_k \in T$ to generate a perturbed input $X' = t_k \circ \ldots \circ t_1(X)$, the corresponding explanation $E'$, and prediction $P'$ by $M$.
We measure the attribution shift score $\mathrm{Score}(E, E')$ using two methods: cosine similarity or $L^\infty$ distance of $E$ and $E'$.
Cosine similarity captures the overall shift of the attribution. $L^\infty$ instead focuses on the most important word.
The attack is successful when
$P=P'$ and $\mathrm{Score}(E, E') > \theta$ where $\theta$ is an application specific threshold parameter.

% An explanation method $S$ is \textit{robust} to an attack $A$ on model $M$ if, for most of the inputs $X_i$, the explanation $E_{io}$ generated for the original sample $X_{io}$ is similar to the explanation $E_{ip}$ generated for the perturbed sample $X_{ip}$ which is created by applying $A$ to $X_{io}$.

% \subsection{Procedure}
% \input{section3/definition}

\subsection{Transformations}
We consider four types of transformations for $T$ to test the adversarial robustness of an explanation method:
misspelling, synonym substitution, word inflection and word deletion.
Each transformation applies necessary constraints to maintain the semantics of the sentence.
% We perturb a sentence using one of these attacks, get attribution scores using explanation method and calculate similarity with the original attributions to measure the robustness.

\subsubsection{Misspellings}
Misspellings refers to replacing the original sentence with a perturbed one such that few of the words are incorrectly spelled due to orthographic changes akin to common human errors in pronunciation or wrongly understood phonetic structure of the word or typographical errors. %The goal of this experiment is to understand whether minor perturbations in input can cause the explanation method to significantly change or remain the same. If a significant change is seen with the same predicted label, we can infer that the explanation method is not robust enough in explaining why the predicted label is justified. For the purpose of this experiment, we pass the original and perturbed sentences into the prediction module and only consider those samples where the predicted labels are the same for both these sentences. If the predicted labels are different, we infer that the model itself isn't robust enough and hence we are unable to comment on the explanation method.
%\todo[inline]{Someone: This whole paragraph seems like repeating what we already said?; Hessel: Probably true and we might be able to reduce this subsection;Shriya: Commented out repeated parts}

\subsubsection{Word Deletion}
We delete a word and verify how the attribution across the perturbed text changes. The attack is again valid only if the meaning of the sentence remains same after deleting a word. 

\subsubsection{Synonym Substitution}
Synonym substitution involves replacing some of the words in the sentence with semantically equivalent words. The replacement words must be chosen such that the semantics of the sentence remain identical or nearly identical to the original sentence. In contrast to the word inflection attack, only synonyms with a different lemma are considered.

\subsubsection{Word Inflection}
The word inflection attack involves replacing words in the sentence with different inflections of the same word. Inflections are words with the same lemma but can have different tense, quantity, etc. 

\subsection{Similarity-based Greedy Search}
\label{subsec:greedy_search}
In order to effectively apply the above transformations, we choose the transformations that generate worst-case adversarial perturbations based on a similarity-based greedy search. The main idea of this algorithm is to find a perturbation that has the highest effect on the attribution similarity score, while maintaining the constraints on how much the sentence can be changed. The constraints we have chosen to apply are as follows:
\begin{enumerate}
    \item The model's prediction for the sentence's label must not change.
    \item A maximum threshold is set for the fraction of words per sentence that can be transformed. (Generally 30\% of the words)
\end{enumerate}

Given a transformation and a candidate word, the algorithm performs the transformation on the word and recomputes the attribution score distribution. This process is repeated until no further words can be changed without violating the constraints or the attribution similarity score drops below the set threshold $\theta$. The sentence obtained as a result of this process is the final perturbed sentence $X'$ for a given transformation. This method can also be used for combined attacks, e.g. a different attack upon every iteration but that combination is out of the scope of this paper.

%\todo[inline]{Hessel: This section need to be reformulated a bit more it seems.; Shriya: Reformulated some of the explanation for clarity; H: Looks good!}

\subsection{Semantic Similarity} \label{sec:sbert}

To further ensure that the meaning of the sentences has not been changed after applying the transformations, we measure the semantic similarity between the original and sentence $X$ and perturbed sentence $X'$. We use the sentence embeddings generated by S-BERT \cite{reimers2019sentence} for this purpose.
\section{Experiments}
\subsection{Dataset}

We run experiments on the Stanford Sentiment Treebank 2 (SST-2) which is a binary sentiment classification dataset released as part of the GLUE Benchmark~\cite{wang2018glue}. It consists of extracted sentences from movie reviews (not the whole review) and either a positive or negative label assigned by a human annotator. For our experiments, we use the validation split which has 872 examples.

\subsection{Model}

The model we used is a RoBERTa base model finetuned on SST-2 \footnote{https://huggingface.co/textattack/roberta-base-SST-2}. The model was publicly shared on the HuggingFace model hub by TextAttack \cite{TextAttack}. The model has an accuracy of 94.04\% on the validation data. 
% It was finetuned for 3 epochs with the Adam optimizer and a learning rate of 2e-05.

\subsection{Attack Implementation}
\subsubsection{Misspellings}
%For misspelling perturbation, we take a given sentence and pass it through the captum explanation framework to get the attribution scores of each word in the sentence. The attribution score is positive or negative based on how much it positively or negatively affects the prediction made.
To generate perturbations of the original sentence for the misspelling transformation, 
we take the most attributed word (positive) and replace if with a misspelled word. 
For the task of finding the misspelled word, we use the ``birkbeck'' dataset~\cite{mitton1985corpora}
which contains misspellings of 6000+ commonly used words. 
If the given word is not present in the dataset, we perform a QWERTY substitution. 
The QWERTY substitution attack is provided by the TextAttack library and 
contains common keyboard based human errors. 
%By performing these operations, we now get the perturbed sentence which is again passed through the same prediction module to obtain the explanation vectors. We now compute the cosine similarity between the original and perturbed explanation vectors and use that as a measure to test robustness of the explanation. %\todo{I think this implementation detail should be moved to experiments as well as some of this stuff is repetitive} 

Table~\ref{tab:misspelling-example-table} shows examples of a few misspelling perturbations. The words that are perturbed are shown in italics.

\begin{table*}[ht]
   \caption{Examples of misspellings attack with the misspelled word in \emph{italics}.}
    \label{tab:misspelling-example-table}
    \centering
    \begin{tabularx}{\textwidth}{lXX}
    \toprule
     & Original Sentence & Perturbed Sentence  \\
      \midrule
    1 & the acting , costumes , music , cinematography and sound are all \textit{astounding} given the production 's austere locales . & the acting , costumes , music , cinematography and sound are all \textit{astoumding} given the production 's austere locales . \\
    2 & a sequence of ridiculous shoot - 'em - \textit{up scenes} .  & a sequence of ridiculous shoot - 'em - \textit{jp seens} .  \\
    \bottomrule
    \end{tabularx}
\end{table*}

\subsubsection{Word Deletion}

%We first calculated attributions of each word in original sentence and deleted word with least attribution score. 
We choose the least attributed words as candidates for word deletion. Since the sentence is tokenized into sub-words, the explanation methods assigns attribution scores to each token rather than the word. So, to calculate the score at the word level, we averaged the scores of all the tokens present in the word. As outlined in Section~\ref{sec:sbert}, we use S-BERT to ensure that the deletion does not significantly change the semantics of the sentence which is very important with this transformation.
%To evaluate the attack, we calculate the cosine similarity of the attribution scores. If the attack did not affect the explanation, the cosine similarity should be high. We manually evaluated few results and chose 0.5 as the threshold. If the similarity score is less than 0.5, we consider the attack as success. Later we used greedy search algorithm which automatically deletes more than one word in a longer sentence.
%\par It is very important to make sure that the meaning of the sentence does not change after perturbation. So, we used sentence bert to calculate the similarity of the original sentence and perturbed sentence. We calculate the embedding of the both the sentences and calculate cosine similarity between them to make sure the sentence did not change much after perturbation. If the cosine similarity is very low, we consider the attack as failure.

\subsubsection{Synonym Substitution}

\begin{table*}[ht]
    \caption{Examples of synonym substitution attack based on WordNet and word embeddings.}
    \label{tab:synonym_examples}
    \centering
    \begin{tabularx}{\linewidth}{lXX}
    \toprule
      &Original Sentence & Perturbed Sentence  \\
      \midrule 
      &\textbf{WordNet}   & \\
       1.1&Or doing \textit{last} year's taxes with your ex-wife & Or doing \textit{endure} year's taxes with your ex-wife \\
       1.2&Unflinchingly \textit{bleak} and desperate & Unflinchingly \textit{stark} and desperate \\
     %& \textbf{BERT} & \\
     %2.1 & & \\
     & \textbf{Embeddings} & \\
      2.1&a \textit{gorgeous} , high-spirited musical from india that exquisitely blends music , dance , song , and high drama . & a \textit{resplendent} , high-spirited musical from india that exquisitely blends music , dance , song , and high drama . \\
      \bottomrule
    \end{tabularx}
\end{table*}

 Three different approaches were attempted for synonym substitution. The first approach involved choosing synonyms for the word using WordNet \cite{fellbaum2010wordnet}. This turned out to be unsuitable for the task as WordNet does not perform word-sense disambiguation. Some of the substitutions produced were of good quality and retained semantic similarity (example 1.2 in Table \ref{tab:synonym_examples}) but others did not. In example 1.1 in Table \ref{tab:synonym_examples}, the original sentence uses the word \emph{last} to mean \emph{previous} but the substitution is the word \emph{endure} which is a different sense of  \emph{last}.

The second approach considered used BERT \cite{devlin2018bert}. As BERT's original training task is predicting masked out words from sentences, this can be adapted to generating substitutions. The word to be substituted is masked out and the masked sentence is passed to the model. The model's top predictions for the masked out word can be used as substitutes in the perturbed sentence. However, since the original word is masked out, BERT only looks at the semantics of the sentence and not the word itself. This leads to it often producing substitutions that are completely unrelated to the original word although they fit in the context of the sentence.

Finally, we perform synonym substitution using counter-fitted word embeddings \cite{mrkvsic2016counter}. Unlike traditional word embeddings, these embeddings are trained with linguistic constraints to ensure that antonyms are not nearest neighbors. This method produced appropriate results for most of the examples and was the final choice.

\subsubsection{Word Inflection}

Word inflection was performed using the LemmInflect\footnote{https://github.com/bjascob/LemmInflect} library. LemmInflect uses a dictionary approach to lemmatize English words and inflect them. It works with out-of-vocabulary (OOV) words by applying neural network techniques to classify word forms and choose the appropriate morphing rules. LemmInflect has a 95.6\% accuracy on the AGID database \footnote{http://wordlist.aspell.net/other/}

\subsection{Results}
\begin{table*}[tb!]
    \caption{Percentage of test samples with changes in the predicted label and semantics of the input sample along with average number of words perturbed and success rate for each type of attack. }
    \label{tab:results-summary}
    \centering
    \begin{tabular}{lcccc}\toprule
    Attack & 
$\Delta$ Label (\%) &
$\Delta$ Semantics (\%) &
$\varnothing$ Perturbations &
Success (\%) \\ \midrule
Word Deletion &
5.5 &
5.0 &
3.0 &
32.1
\\
Synonym Substitution &
9.0 &
5.6 &
1.6 &
67.1 \\
Inflection &
1.2 &
1.3 &
2.0 &
39.5 \\
Misspelling &
4.3 &
8.5 &
1.8 &
86.0 \\
\bottomrule
    \end{tabular}
\end{table*}

\begin{figure}[t!]
    % \begin{subfigure}[b]{0.4\linewidth}
        \centering
        \includegraphics[width=0.83\textwidth]{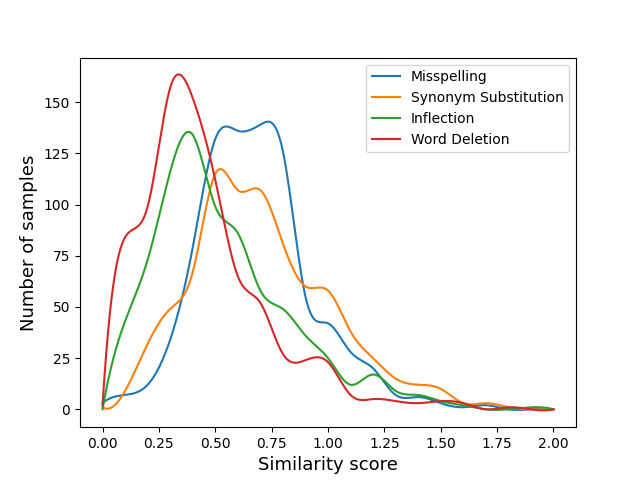}
        % \caption{$L_\infty $ distance}
        \caption{Number of test samples v. $L_\infty $ distance between explanation vectors before and after adversarial perturbation to the input.}
        \label{fig:l-inf-plot}
    % \end{subfigure}
    % \caption{Number of test samples v/s similarity/distance between explanation vectors before and after adversarial perturbation to the input.}
    % \label{fig:similarity-line-plot}
\end{figure}

Table \ref{tab:results-summary} shows summary of the various attacks.
Figure~\ref{fig:cosine-line-plot} shows the number of successful attacks for each bucket of cosine similarity between the explanation vectors before and after an adversarial perturbation. Figure~\ref{fig:l-inf-plot} shows the number of successful attacks for each bucket of $L_\infty$ distance between the explanation vectors before and after an adversarial perturbation.
% From the figure, we can infer that generally, the misspelling attack has been successful. 
An attack is referred to as being \emph{successful} if
the prediction does not change,
\emph{and} the explanation vectors differ significantly for the original and the perturbed sentence \ie, having low cosine similarity.

\subsubsection{Misspellings}
\begin{figure}[t!]
    \centering
    % \begin{subfigure}[b]{0.4\linewidth}
        \centering
        \includegraphics[width=0.83\textwidth]{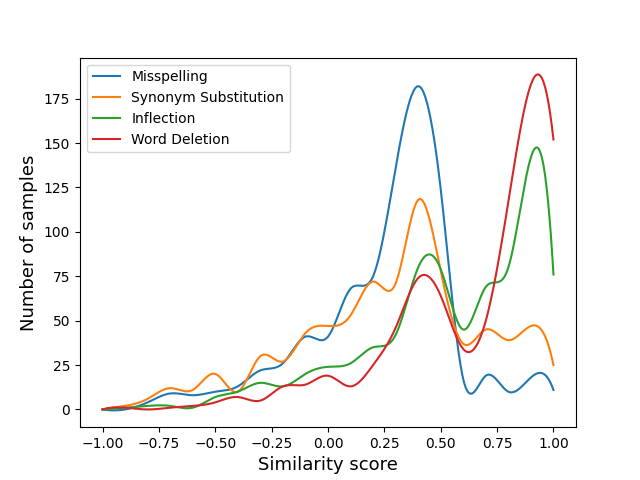}
        % \caption{Cosine similarity}
        \caption{Number of test samples v. cosine similarity between explanation vectors before and after adversarial perturbation to the input.}
        \label{fig:cosine-line-plot}
    % \end{subfigure}
\end{figure}

As shown in Figure~\ref{fig:cosine-line-plot}, majority of the samples have 
cosine similarity between explanation vectors around 0.4 for the misspelling attack. 
Around 770 out of 872 samples had explanations with a similarity of less than 
the 0.6 threshold with respect to the original sentences. 
This implies that even though the predicted label was the same, 
the explanation method is not robust of misspelling based perturbation.
A potential reason for this could be 
that since we are misspelling the most attributed word in the original 
sentence, the attribution scores get distributed over other words in the 
perturbed sentence. This may lead to significantly 
different explanation vector than the original and lead to low cosine similarity
between explanation vectors before and after the attack.

%Figure \ref{fig:cosine-line-plot} and \ref{fig:l-inf-plot} shows number of sentences vs similarity score between attributions of original and perturbed sentences. We can observe that most of the samples lie below 0.5 similarity score for misspelling attack.

\begin{figure}[!tb]
    \centering
    \subfloat[Misspelling\label{fig:misspellings-results-sent-sim}]{%
    \includegraphics[width=0.45\textwidth]{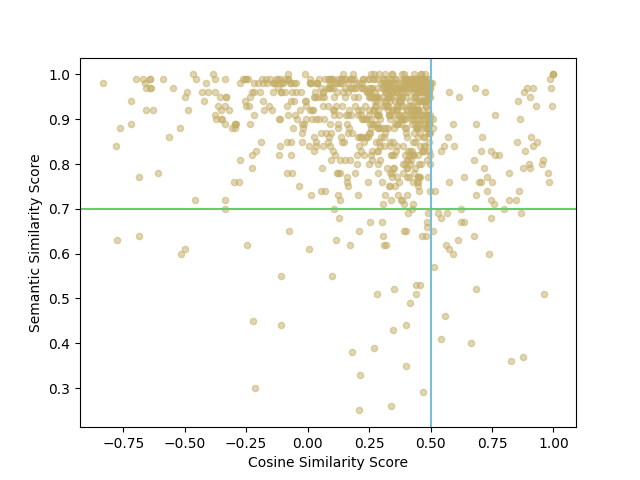}
    }\hfil
    \subfloat[Word Deletion\label{fig:deletion-results-sent-sim}]{%
    \includegraphics[width=0.45\textwidth]{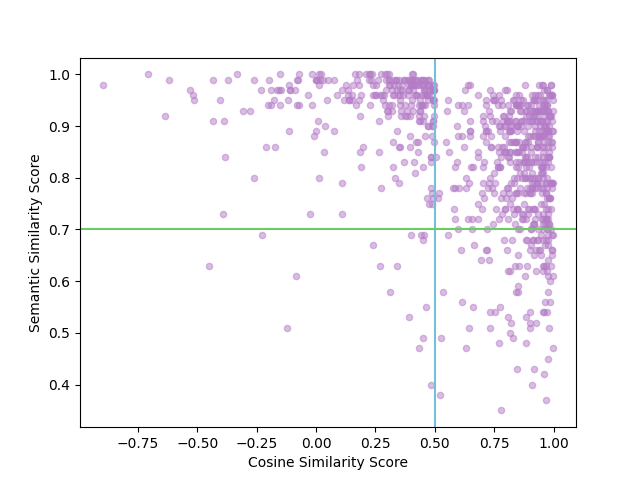}
    } \\
    \subfloat[Synonym Substitution\label{fig:synonym_sentence_similarity}]{%
    \includegraphics[width=0.45\textwidth]{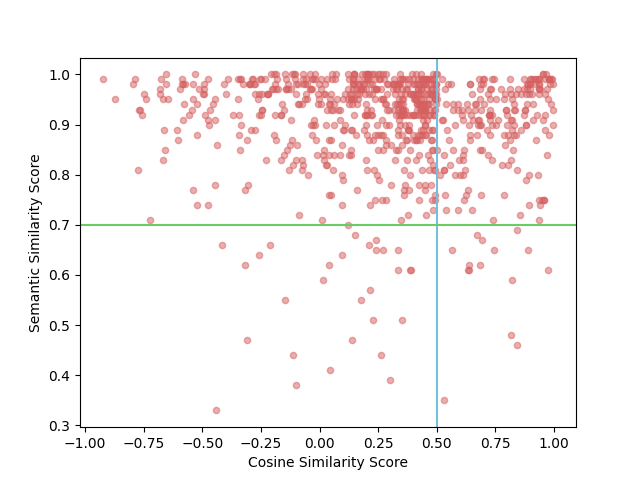}
    }\hfil
    \subfloat[Word Inflection\label{fig:inflection_sentence_similarity}]{%
    \includegraphics[width=0.45\textwidth]{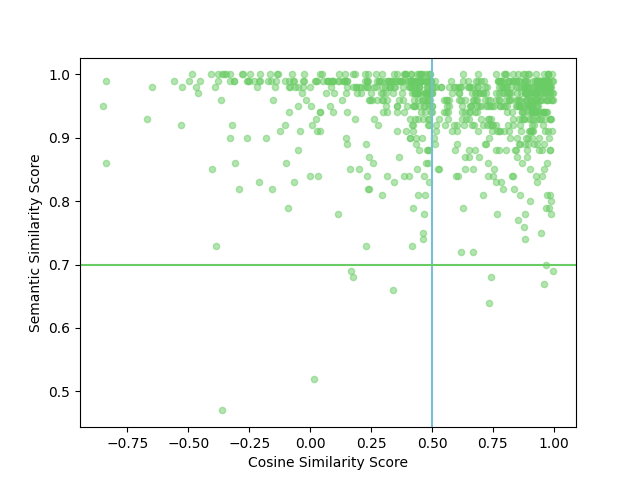}
    }
    \caption{Semantic similarity versus cosine similarity of explanation vectors of original and perturbed input samples.}
    \label{fig:results}

\end{figure}

It is also important to check whether or not the perturbations themselves 
significantly change the semantics of the original sentences which can cause the explanation method to fail. 
Figure~\ref{fig:misspellings-results-sent-sim} shows scatter plot of BERT-based semantic
similarity between original and perturbed sentences 
and cosine similarity between corresponding explanation vectors.
A successful attack would have most datapoints in the top-left corner which correspond to high semantic similarity between the original and perturbed sentence along with low cosine similarity between corresponding explanations. For misspellings attack, 705 examples out of the total 872 examples have sentence similarity greater than 0.7 and cosine similarity less than 0.5.

%\begin{figure}[tb]
%    \includegraphics[width=\linewidth]{figs/misspelling-plot-cumulative.png}
%    \caption{Results of Misspellings Attack (Cumulative)}
%    \label{fig:misspellings-results-cumulative}
%\end{figure}
%\todo{Should we remove this figure? Considering all other plots are not cumulative}

\subsubsection{Word Deletion}
%\begin{figure}[tb]
%    \centering
%    \includegraphics[width=0.8\linewidth]{figs/word-deletion-result-1.png}
%    \caption{Word Deletion - 1 word deleted}
%    \label{fig:word-deletion-result-1}
%\end{figure}
%\begin{figure}[tb]
%    \centering
%    \includegraphics[width=0.8\linewidth]{figs/word-deletion-result-2.png}
%    \caption{Word Deletion - 2 words deleted}
%    \label{fig:word-deletion-result-2}
%\end{figure}
% We used 0.5 threshold for similarity score to evaluate the word deletion attack. 

Figure~\ref{fig:cosine-line-plot} shows that majority of the samples have high cosine similarity
between explanation vectors before and after deleting the least important word from the sentence.
This is apparently because we are deleting \emph{least} important word in sentence 
according to the attribution scores, and this causes marginal changes to overall attribution 
scores for the sentence.
We experimented with deleting one or two least important words in a single attack as well as applying greedy search with threshold of 0.5 as explained in Section~\ref{subsec:greedy_search}. The attack was successful 
for only 7\% of the sentences for single word deletion and the attack was successful for 65 sentences out of 755 when deleting two words. 
% As the length of the sentence varies, we can vary the number of words perturbed as well. 
With greedy search, the attack was successful for 32\% of the samples. Overall, Integrated gradients explanation method is largely robust against our proposed word deletion attack.

\subsubsection{Synonym Substitution}

The synonym substitution attack automated with greedy search produced similar results to misspelling although it was slightly less effective. In Table \ref{tab:results-summary} we can see that it has the second highest percentage of succesful attacks at 67\%. In Figure \ref{fig:cosine-line-plot}, it can be observed that the synonym substitution peak is slightly to the right of the misspelling peak, indicating slightly higher cosine similarity in the attribution vectors. However, it remains much to the left of the inflection and word deletion peaks. In Figure \ref{fig:synonym_sentence_similarity} we can also note that the number of points in the top left corner is lower than for the misspelling plot but higher than for the remaining transformations. 

\subsubsection{Word Inflection}

The word inflection attack is only slightly more effective than the word deletion one. It has a success percentage of 39.44. Unsurprisingly, most of the points in Figure \ref{fig:inflection_sentence_similarity} are well above the 0.7 sentence similarity line as this attack produces only a very minor change in each word affected. However, it is also not very effective at changing the attribution vectors, possibly for the same reason.

\section{Conclusion}
Explanation methods are crucial going forward into the future where ML models will be deployed in critical applications. The goal of our work is to devise methods to automatically test robustness of the explanation methods specifically in text domain. We performed multiple types of attacks on text under constraints that ensured meaning remained consistent and devised methods to measure the change in explanations. We primarily focused on Integrated Gradients explanation method but our work could easily be extended to other methods as well. We found Integrated Gradients method is not robust against Misspelling and Synonym substitution attacks as the explanation changed heavily upon attack. We believe this is a small step in the right direction towards testing robustness of explanation methods.

\par
%We described few shortcomings with our approaches in the experiments section. We plan to look into alternative methods to find least important word in a sentence in word deletion experiment. We also plan to explore different approaches to generate better perturbed sentences in synonym substitution experiment. We found Integrated Gradients explanation method is not very robust to misspelling attack. But we also observed that model mispredicted nearly 15\% of the sentences against misspelling attack. The model could be not very robust to the misspelling attawck. It is possible the model itself is responsible for change in attribution scores instead of the explanation method. To ensure our observations, we could test our attacks on a robust model trained on adversarial examples.

%We also plan to explore other similarity measures like Area Over Perturbation Curve (AOPC) (\cite{aupc-samek}). 

%We attacked explanation method with different types of perturbations individually. We are curious to examine the robustness by simultaneously perturbing a sentence using multiple methods.\todo[inline]{Hessel: Can we remove future work file? Than only remaining task is to rewrite the first paragraph of Conlcusion a bit I guess?}

\bibliographystyle{splncs04}
\bibliography{bibliography}

\end{document}